\documentclass{article}

\usepackage[final, nonatbib]{neurips_2022}

\usepackage[utf8]{inputenc} % allow utf-8 input
\usepackage[T1]{fontenc}    % use 8-bit T1 fonts
\usepackage[hidelinks]{hyperref}       % hyperlinks
\usepackage{url}            % simple URL typesetting
\usepackage{booktabs}       % professional-quality tables
\usepackage{amsfonts}       % blackboard math symbols
\usepackage{nicefrac}       % compact symbols for 1/2, etc.
\usepackage{microtype}      % microtypography
\usepackage{xcolor}         % colors
\usepackage{graphicx}
\graphicspath{ {./images/} }
\usepackage{amsmath}
\usepackage{hyperref}

\usepackage[
backend=biber,
style=numeric,
sorting=none,
language = american
]{biblatex} %Imports biblatex package
\title{Dynamic Survival Transformers for Causal Inference with Electronic Health Records}

\author{%
  Prayag Chatha \\
  Department of Statistics \\
  University of Michigan \\
  \texttt{pchatha@umich.edu} \\
   \And
  Yixin Wang \\
   Department of Statistics \\
   University of Michigan \\
   \AND
  Zhenke Wu \\
   Department of Biostatistics \\
   University of Michigan \\
   \And
   Jeffrey Regier \\
   Department of Statistics\\
   University of Michigan\\
}

\usepackage{todonotes}

 \newcommand{\ind}{\perp\!\!\!\!\perp}

\makeatletter
\renewcommand{\@noticestring}{Accepted to the NeurIPS 2022 Workshop on Learning from Time Series for Health.}

\makeatother

\begin{document}

\maketitle

\begin{abstract}
    In medicine, researchers often seek to infer the effects of a given treatment on patients' outcomes. However, the standard methods for causal survival analysis make simplistic assumptions about the data-generating process and cannot capture complex interactions among patient covariates. We introduce the Dynamic Survival Transformer (DynST), a deep survival model that trains on electronic health records (EHRs). Unlike previous transformers used in survival analysis, DynST can make use of time-varying information to predict evolving survival probabilities. We derive a semi-synthetic EHR dataset from MIMIC-III to show that DynST can accurately estimate the causal effect of a treatment intervention on restricted mean survival time (RMST). We demonstrate that DynST achieves better predictive and causal estimation than two alternative models.
\end{abstract}

\section{Introduction}

Medical practitioners are often interested in the effect of a treatment on a patient's survival time until an event of interest. For instance, if a patient is prescribed a certain antibiotic, how will that affect their risk of experiencing sepsis in the next 24 hours? The field of causal survival analysis is concerned with estimating treatment effects on time-to-event outcomes given incomplete (censored) data; classical techniques such as the Kaplan-Meier curves \cite{km} and the Cox regression model \cite{cox} are extensively used despite their limitations. Kaplan-Meier curves are a descriptive tool that do not model individual survival trajectories, while the Cox model assumes proportionality of hazard functions, which may be unrealistic. Meanwhile, the rise of electronic health records (EHRs) has led to an abundance of multi-concept longitudinal data: a setting for \textit{observational} causal inference, if randomized controlled trials prove impractical or unethical.

With this observational setting in mind, we propose the Dynamic Survival Transformer (DynST), a deep-learning survival model that estimates individual survival probabilities over time from EHR data. DynST is built on the Transformer \cite{attn}, a recent neural network architecture that has achieved state-of-the-art results in sequence-to-sequence learning, particularly in NLP \cite{bert}. Transformers can flexibly model individual survival trajectories without making simplifying parametric assumptions about the data-generating process. Unlike previous survival transformers \cite{TDSA, BERTSurv, survtrace} DynST exploits both static and time-varying features to capture how a patient's event risk evolves over time. Several works have applied transformers to prediction problems in EHR data \cite{behrt, medbert, cehr-bert, causal_transformer}, motivated by similarities between EHRs and text, but DynST is the first transformer used to estimate the average effect of a treatment intervention on survival outcomes. Using a semi-synthetic dataset derived from MIMIC-III \cite{mimic3}, we show that DynST can improve on baseline methods in survival time prediction and causal inference.

\section{Problem setup}
We observe survival data taking the form $(X_i, O_i, \delta_i)_{i=1}^n$, where $X_i$ represents the $i$-th patient's features, $O_i$ is the observed (and possibly censored) time to the event, and $\delta_i$ is a binary variable indicating whether the event was observed or not, due to censoring. If $\delta_i = 0,$ the $i$-th patient is right-censored, so the event takes place \textit{after} $O_i.$ Let $T_i$ represent the uncensored survival time and let $C_i$ be the censoring time. Then, $O_i = \min\{T_i, C_i\},$ and $\delta_i = \mathbf{1}(T_i \leq C_i).$ In this paper, we assume conditionally independent censoring, i.e., $T_i \ind C_i \mid X_i.$ We also assume a discrete survival setup, where $T_i \in \{1, 2, \ldots, ...t_\mathrm{max}\}$ and time steps are evenly spaced. The \textit{hazard function} 
\begin{equation}
h(t \mid X) = P_X(T = t \mid T \geq t)
\end{equation}
is the risk of failure at time $t$ given that the patient has survived thus far. The \textit{survival probability} $S$ at time $t$ is
\begin{equation}
S(t \mid X) = P_X(T > t) = \prod_{\tau=1}^t(1 - h(\tau \mid X)).
\end{equation}
The expected survival time is defined as
\begin{equation}
\mathbb{E}[T \mid X] = \sum_{t = 1}^{t_\mathrm{max}} S(t \mid X).
\end{equation}
Lastly, given a cutoff time $\tau,$ the restricted mean survival time (RMST) \cite{rmst, nn-rmst} is defined as
\begin{equation}
Y_\tau = \mathbb{E}_X[\min\{T, \tau\}] = \frac{1}{n} \sum_{i=1}^n \left ( \sum_{t = 1}^{\tau} S(t \mid X_i) \right ).
\end{equation}
RMST can be thought of as the expected survival time up to time $\tau,$ averaged over the population of all patients.

\section{Methods}

\begin{figure}
  \centering
  \includegraphics[scale=0.4]{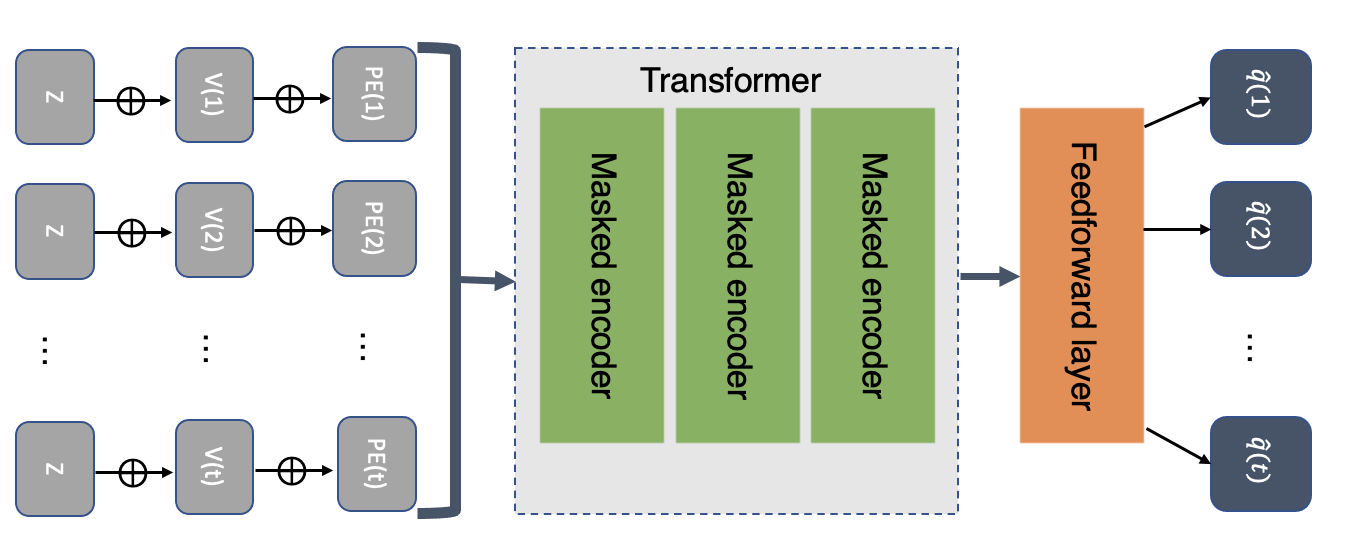}
  \caption{A diagram of DynST modeling the hazard function from a single patient's EHR data.} 
  \label{fig:dynst}
\end{figure}

\subsection{Model architecture}

Let $X$ denote the features of a single patient; we suppress the patient index for readability. $X$ consists of $p$ static features $Z_1, \ldots Z_p,$ collectively denoted as $Z,$ and $q$ time-varying features, $V_1, \ldots V_q,$ collectively denoted as $V.$ Here each feature $V_j$ is a time series vector $(V_j^{(1)}, V_j^{(2)}, \ldots, V_j^{(t_\mathrm{max})}).$ Static variables may include initial diagnoses, whereas a sequence of lab measurements is an example of a time-varying feature. Let $\overline{V^{(t)}} = (V_j^{(1)}, \cdots V_j^{(t)}).$ At each time step $t,$ the Dynamic Survival Transformer models $q(t; Z, \overline{V^{(t)}}) = 1 - h(t \mid Z, \overline{V^{(t)}}).$ That is, DynST predicts the complement of each patient's hazard function using static features and the available history of time-varying features.

Figure \ref{fig:dynst} illustrates DynST's architecture. The model transforms each patient's medical records through the following procedure:

\begin{enumerate}
    \item \textbf{Input embeddings}: A linear layer transforms time-varying input features $V$ into a $t_\mathrm{max} \times d_{model}$ matrix $W_V$. Another linear transformation on static input features $Z$ yields a length $d_{model}$ vector $W_Z$. We add $W_Z$ to each row of $W_V,$ obtaining an embedded input sequence $W = (W_1, \ldots, W_{t_\mathrm{max}}).$
    \item \textbf{Positional encodings} ($P_E$): Following the approach in \cite{attn}, a deterministic sinusoidal function encodes time steps $1, \ldots, t_\mathrm{max}$ as vectors of length $d_{model},$ $P_E(1), \ldots, P_E(t_\mathrm{max}).$  We add these encodings to the corresponding rows of $W.$
    \item \textbf{Autoregressive Transformer Encoder}: $W$ is fed through $m$ transformer encoder layers, which combine multihead self-attention (eight heads) with a feedforward network and dropout. We apply autoregressive masking so that the $t$-th position of $W$ cannot attend to (i.e., cannot learn from) future positions. We end up with the transformed output sequence $\phi(W) = (\phi(W_1), \ldots, \phi(W_{t_\mathrm{max}})).$
    \item \textbf{Hazard function output}: Following \cite{TDSA}, we apply a two-layer feedforward network with a final sigmoid layer to obtain a vector of probabilities $\hat q(1), \ldots, \hat q(t_\mathrm{max}).$ These estimate the complement of the hazard function.
\end{enumerate}

Let $\hat q_i(t)$ denote the estimated hazard function for the $i$-th patient. From the predicted hazards, we estimate patient survival probabilities as $\hat S_i(t) = \prod_{\tau}^{t} \hat q(t ; Z_i, \overline{V_i^{(t)}})$ and expected survival time as $\hat T_i = \sum_{t=1}^{t_\mathrm{max}} \hat S_i(t).$ We note that our prediction $\hat S_i(t)$ has no access to future covariates occurring after time step $t.$

\subsection{Training}

Our model jointly minimizes two objective functions. The first, like the loss functions in \cite{rnn-surv,TDSA}, is an adaption of cross-entropy loss to censored data:
\begin{equation}
    \mathcal{L}^{(i)}_1 = - \left [ \sum_{t=1}^{O_i-1} \log \hat S_i(t) +  \sum_{t=O_i}^{t_\mathrm{max}} \log (1 - \hat S_i(t) \right ] \cdot \delta_i - \left [ \sum_{t=1}^{O_i} \log \hat S(t)\right ] \cdot (1 - \delta_i).
\end{equation}

For censored patients ($\delta_i = 0$), this means maximizing survival probabilities over the period of observation. For uncensored patients, survival probabilities up to the failure time are maximized, and subsequent survival probabilities are minimized. The second loss penalizes the error of the predicted survival times:

\begin{equation} \label{mae}
    \mathcal{L}^{(i)}_2 = \left | O_i - \hat T_i \right |\cdot \delta_i + \max\{0, O_i - \hat T_i\} \cdot (1 - \delta_i).
\end{equation}
That is, when failure time is observed, the absolute error of $\hat T_i$ adds to the loss. For censored observations, loss is incurred only when the estimated survival time is before the time of censoring. The losses are combined and summed over all patients:
\begin{equation}
    \mathcal{L} = \sum_{i=1}^n (1 - \alpha) \mathcal{L}^{(i)}_1 + \alpha \mathcal{L}^{(i)}_2,
\end{equation}
with $\alpha$ a tuning hyperparameter. We used the Adam optimizer with weight decay and trained DynST in minibatches on an NVIDIA GTX 2080 Ti GPU.

\section{Experiments}

\subsection{Semi-synthetic dataset}

We derived a semi-synthetic longitudinal survival dataset from MIMIC-III, a freely accessible database of de-identified EHRs from intensive care stays \cite{mimic3,mimicpaper}. After discarding records for stays shorter than 16 hours in length, we obtained demographic information, diagnoses, and hourly lab results for 30,323 patients. For a synthetic outcome event simulating in-hospital infection, we generated patient survival trajectories based on a subset of static and time-varying features as well as a predetermined treatment effect. After truncating patient histories longer than 128 hours, 39\% of all patients had censored outcomes. For further details of the data simulation process, see Appendix A.

\subsection{Predictive survival analysis}
To highlight DynST's advantages as a component in causal inference, we demonstrate its improved performance in individualized survival prediction compared to two baselines: a Cox Oracle model and a survival transformer using only static features (Static ST). We compare how well these models estimate patient survival times in terms of mean absolute error (MAE). Table \ref{predict} reports model performances on held-out test data, averaged over six trials. We used a 70/15/15 random split into training, validation and test sets; hyperparameter spaces and other training details are given in Appendix C. Both transformer-based models achieve a lower MAE than the Cox model, which assumes a misspecified model of the hazards. DynST shows a small but significant improvement over its static counterpart, indicating that DynST can capture the effect of time-varying features on patients' survival probabilities.

\begin{table}
  \caption{Performance of DynST and baselines in predicting patient survival times (mean $\pm$ SD)}
  \label{predict}
  \centering
  \begin{tabular}{lll}
    \toprule
    Model     & Mean Abs. Error \\
    \midrule
    Cox Oracle & $16.04 \pm 0.25$     \\
    Static ST     &  $11.42 \pm 0.23$     \\
    DynST (ours)     & $\mathbf{11.19 \pm 0.24}$ \\
    \bottomrule
  \end{tabular}
\end{table}

\subsection{Causal survival analysis}

\begin{table}
  \caption{Comparison of estimators of average treatment effect on RMST (average bias $\pm$ SD)}
  \label{causal}
  \centering
  \begin{tabular}{llll}
    \toprule
    Method     & $\tau = 8 $    & $ \tau = 12$ & $\tau = 16$ \\
    \midrule
    Unadjusted Difference in Means & $-0.502 \pm 0$ & $-1.11 \pm 0$ & $-1.88 \pm 0$ \\
    Cox Regression & $ -0.260 \pm 0.0010$  & $-0.523 \pm 0.0012 $ & $-0.750 \pm  0.0080 $  \\
    Logistic IPW    & $0.257 \pm 0$ & $0.35 \pm 0$ & $0.416 \pm 0$ \\
    DynST Regression     & $-0.146  \pm 0.041$  & $-0.160 \pm 0.078 $ & $-0.190 \pm 0.12$ \\
    AIPW (DynST + Logistic) & $ \mathbf{-0.0390 \pm 0.013}$ & $\mathbf{0.0131 \pm 0.056}$ & $\mathbf{0.127 \pm 0.082}$ \\
    \bottomrule
  \end{tabular}
\end{table}
We treat restricted mean survival time (RMST) as the outcome variable. Given a binary treatment variable $A$, our causal estimand of interest is the \textit{average treatment effect} (ATE) on RMST,
\begin{equation}
    \psi = \mathbb{E}[Y_\tau(1) - Y_\tau(0)].
\end{equation}
ATE is the expected difference between potential outcomes under treatment and control. See Appendix B for a review of the potential outcomes framework for causal inference.

In Table \ref{causal}, we compare several methods of estimating of ATE in terms of bias, averaged over six trials. We compared (1) the unadjusted difference in means between treated and control groups, (2) a Cox model outcome regression (OR), (3) a inverse-propensity weighted (IPW) estimator using a logistic model of propensity scores, (4) OR using DynST, and (5) an augmented IPW (AIPW) estimator combining (3) and (4). For details of how these models were fitted, see Appendix D. The two methods that used DynST to model patient outcomes achieved the best estimation of ATE. This suggests that DynST can successfully adjust for confounding variables. The Cox model and Logistic IPW performed worse; they assume misspecified models of response and treatment.
\section{Discussion}

DynST can model patient survival under complex, time-varying interactions among variables. Combining DynST's flexible outcome modeling with knowledge of the treatment assignment mechanism can result in unbiased estimation of treatment effects. Cross-fitting \cite{doubleML} may help with selecting optimal hyperparameters for causal inference. As future work, we can apply DynST to quantify the real-world causal effects of treatments such as COVID-19 vaccines, for which there is ample EHR data.  We can also generalize the model to handle the effects of time-varying treatments and outcomes. Lastly, we can experiment with further tailoring the training procedure of DynST toward causal inference by jointly modeling outcome response and treatment assignment \cite{dragonnet}.

\printbibliography

\appendix

\section{Semi-synthetic EHR dataset}

To preprocess the raw Medical Information Mart for Intensive Care (MIMIC-III) database, we follow the MIMIC-Extract pipeline \cite{mimic-extract}. Since time-varying features are reported in hourly intervals, in our experiments, each time step corresponds to one hour. We discard patients who have fewer than 16 hours of data and truncate all patient records longer than 128 hours in length. In addition, we scale and center all continuous variables to have mean zero and a standard deviation of one. As estimates of ground-truth causal effects are generally unverifiable, we build a semi-synthetic dataset in which the treatment assignment and outcome response are simulated from real observed features. 

First, we develop an assignment mechanism for the synthetic treatment variable $A$. We assume that this treatment represents an intervention that occurs at the very start of a patient's observed history. We define an indicator variable $Z_*$ denoting whether a patient is ``severely ill.'' We consider three conditions, \textit{hypertension}, \textit{coronary atherosclerosis}, and \textit{atrial fibrillation}. If a patient has been diagnosed with more than one of these illnesses, then $Z_* = 1$; otherwise, $Z_* = 0.$ The propensity score is then
\begin{equation}
    P(A = 1 \mid X) \equiv \pi(X_i) = \begin{cases}
    0.8, & \text{if $Z_* = 1$} \\
    0.2, & \text{if $Z_* = 0$}.
    \end{cases}
\end{equation}
Every patient has a chance of being in either the treatment or the control arm, though this chance depends on information that will also influence their time-to-event outcomes.

Next, we simulate patient survival trajectories. Rather than directly modeling a patient's survival time, we model the hazard function at each hourly time step. This way, we ensure that survival probabilities adapt dynamically with patient medical histories. To generate the hazards, we use a subset of the available features: five static and four time-varying. One is the synthetic treatment; the rest were selected because they correlate with treatment. Thus we can induce a confounding structure between treatment and outcome. The synthetic hazard function is defined as follows:

\begin{equation}
    h(t) = \underset{(1)}{H_0 \exp(- \lambda t)} \cdot \underset{(2)}{\exp(\theta A)} \cdot \underset{(3)}{\exp(\sum_{j=1}^4 \beta_j Z_j)} \cdot \underset{(4)}{\exp(\log(1.02) t Z_*)} \cdot \underset{(5)}{\exp(\sum_{j=1}^4 \gamma_j g(V_j^{(t)}))}.
\end{equation}

\begin{enumerate}
    \item \textbf{Baseline hazard}: We model the baseline hazard function with exponential decay, setting $H_0 = 0.001$ and $\lambda = 0.25.$ All else being equal, a patient's risk decreases the longer they survive (a.k.a. the Lindy effect). 
    \item \textbf{Treatment effect}: $A$ is the binary treatment variable, and $\theta = -0.5$ parameterizes the treatment effect. A treatment intervention will decrease a patient's lifetime risk.
    \item \textbf{Static variables}: We include the linear effects of four binary, static variables $Z_1, \ldots, Z_4$ along the lines of a Cox regression model. The first variable is whether a patient is male; the remaining three are the presence of ICD-9 codes for hypertension, coronary atherosclerosis, and atrial fibrillation. We generated the coefficients $\beta_j$ from a $\mbox{Uniform}(0.7, 1.2)$ distribution.
    \item \textbf{Temporal interaction term}: $Z_*$ is the indicator variable for severely ill patients (those with two or more of the above diagnoses present). This variable has a temporal interaction, thus violating the proportional hazards assumption \cite{Austin2017}. When $Z_*$ is 1, the hazard increases by two percent at every time step, modeling a severely ill patient's deteriorating health.
    \item \textbf{Dynamic variables}: Our time-varying features $V_1, \ldots, V_4$ correspond to vital readings for hematocrit, hemoglobin, platelets, and mean blood pressure. Coefficients $\gamma_j$ are generated from a $\mbox{Uniform}(0.1, 0.3)$ distribution. We set
    \begin{equation}
    g(V_j^(t)) = \begin{cases} 0, & \text{if $V_j^{(t)} \geq 0$} \\ \max\{(V_j^{(t)})^2, 3\}, & \text{else.} \end{cases}
    \end{equation}
    Thus, whenever $V_j$ drops below the average (i.e., zero) value, it makes a quadratic contribution to the log hazard. We used clipping to avoid inflated hazards.
\end{enumerate}

We constrain hazards to lie in the range $(10e^{-8}, 0.1)$ for reasons of stability. At each time step $t,$ we calculate the survival probability to be $S(t) = \prod_{\tau=1}^t(1 - h(t)).$ To introduce a degree of unexplained randomness to the system, we add Gaussian noise ($\sigma=0.5$) to the logits of the survival probabilities. Lastly, we simulate patients' time-to-event trajectories with the modeled survival probabilities. At each time step $t =1, \ldots, t_\mathrm{max}, $ we use a patient's survival probability to generate a Bernoulli random variable. The first time step that results in a zero-value we take to be a patient's survival time. If no zeros were generated at any $t,$ then the patient is right-censored. In the resulting dataset, 39\% of patients have right-censored survival times. The average time to censoring or failure is 27.8 hours. 

To derive the ``true'' ATE on RMST, we create two copies of the MIMIC dataset, one where every patient has $A_i = 1$ and the other where every patient has $A_i = 0.$ Let $Y_{i, \tau}(A_i, X_i)$ denote the simulated RMST at cutoff $\tau$ for patient $i.$ We calculate true ATE as
\begin{equation}
    \psi \approx \frac{1}{n} \sum_{i=1}^n \left ( Y_{i, \tau}(1, X_i) - Y_{i, \tau}(0, X_i) \right ).
\end{equation}
 Since $n = 30323,$ this should be an accurate approximation of the population ATE.

\section{Causal inference}

We follow the potential outcomes (Neyman-Rubin) framework for causal inference. We observe data of the form $(X_i, A_i, Y_i)$ for subjects $i=1, \ldots, n,$ where $X_i$ denotes subject covariates, $A_i$ denotes a binary treatment indicator, and $Y_i$ is an outcome of interest. The \textit{potential outcome} of $Y_i$ under an intervention $A_i = a$ (i.e., assigning a subject to patient or control) is written as $Y_i(a).$ The causal relationship between $A$ and $Y$ is confounded if $\mathbb{E}[Y(a)] \neq \mathbb{E}[Y \mid A = a].$ That is, certain confounding variables predispose subjects who are observed in one treatment arm toward certain outcomes. This does occur in randomized controlled trials (RCTs), but in observational causal inference, the goal is to adjust for confounding variables to identify the true cause-and-effect relationship between $A$ and $Y.$ We may identify causal effects in an observational study if the following assumptions hold.
\begin{enumerate}
    \item \textbf{Consistency}: The potential outcome for a patient assigned intervention $A = a$ is the same as the outcome for a patient observed ``in the wild'' with treatment $A = a.$ That is, $A_i = a \implies Y_i(a) = Y_i.$
    \item \textbf{Positivity}: All subjects have a non-zero chance of being in the treatment or control groups. Let $\pi(X_i) = P(A_i = 1 \mid X_i)$ be the \textit{propensity score} for the $i$-th subject. We then assume that $\pi(X_i) \in (0, 1)$ for all $i.$
    \item \textbf{Exchangeability}: Also referred to as ignorability or no unobserved confounders, we assume that potential outcomes are independent of observed treatment status when conditioning on the set of covariates $X.$ Formally, we assume that $Y(a) \ind A \mid X.$ For subjects with comparable features, an observational study emulates an RCT. 
\end{enumerate}
Under these assumptions, $E[Y(a) \mid X] = E[Y \mid A=a, X],$ allowing us to infer potential outcomes from observed data. Still, a key difficulty is that only one of two potential outcomes is observed per subject. (If a patient was treated, we do not observe their potential outcome under control assignment.) The other is a \textit{counterfactual} outcome. The potential outcomes framework is akin to reformulating causal inference as a missing data problem. In this paper, we assume all of the above assumptions. In non-synthetic datasets, only the second assumption is generally testable. For a textbook-level treatment of the potential outcomes framework, we refer the reader to \cite{whatif}.

A common causal estimand is \textit{average treatment effect}, defined as
\begin{equation}
    \psi = \mathbb{E}_X[Y(1) - Y(0)],
\end{equation}
for some outcome variable $Y.$ This quantifies the average difference in outcomes between two hypothetical populations, one in which all subjects were assigned treatment and another in which everyone was made a control. In general, due to confounding, 
\begin{equation}
    E[Y \mid X = 1] - E[Y \mid X = 0] \neq \psi.
\end{equation}
We refer to the observed difference in outcomes between treated and control groups as the \textit{unadjusted difference in means}.

Two common approaches to estimating ATE are inverse propensity weighing (IPW) and outcome regression. An IPW estimator involves fitting a model (e.g., a logistic regression) that estimates the treatment probabilities $\pi(X)$ and using these propensity scores to create a pseudo-population in which all individuals are equally likely to receive treatment or not. Outcome regression (a.k.a. standardization, g-computation) requires estimating the conditional expectation $E[Y \mid A, X]$ to determine the effect of $A$ on $Y$ while adjusting for potential confounders $X$. Synthesizing these methods, the Augmented IPW (AIPW) estimator fits nuisance models for both treatment assignment and outcome response. The AIPW has  the desirable double robustness property, whereby it gives an unbiased estimate of ATE if either of its components are unbiased. Furthermore, even when one or both of the nuisance models have a relatively slow rate of convergence (e.g., neural networks), under certain conditions, the AIPW can achieve the linear semiparametric efficiency bound \cite{doubleML}. For an overview of techniques for estimating ATE and double robustness, see \cite{aipw}.

\section{Prediction experiment}

We compare how well DynST, a static survival transformer (i.e., one with no access to time-varying features), and an oracle Cox model (i.e, one that uses only the features relevant to the true hazard function) make predictions of patients' expected survival times. We measure this performance in terms of mean absolute error. We define our MAE as follows:
\begin{equation}
    \mathcal{C}_{\mathrm{MAE}} = \frac{1}{n} \sum_{i=1}^n \left [ \left | O_i - \hat T_i \right |\cdot \delta_i + \max\{0, O_i - \hat T_i\} \cdot (1 - \delta_i) \right ].
\end{equation}
Similar to the loss function in equation (4), this metric accounts for patients with observed and censored survival times alike. When a patient's time-to-event is observed, the absolute difference between estimated and true survival times contributes to error. When a patient's survival is censored, an estimated survival time that falls short of the censoring time counts as error.

We used the following hyperparameter space to train DynST and the static transformer:
\begin{itemize}
    \item Latent dimension $d_{model} \in \{32, 48, 64\}$
    \item Number of transformer blocks $m \in {2,3,4}$
    \item Batch size $b \in \{16, 32\}$
    \item Joint loss ratio $\alpha \in \{0, 0.1, 0.2\}$
    \item Number of epochs $k_{epoch} \in \{1,2,3,4,5\}$
\end{itemize}
We used MAE as the criterion for early stopping and fixed the dropout proportion at 0.1. Using a 70/15/15 ratio to randomly split the data into training, validation, and test sets, we selected optimal hyperparameters using the validation set and reported MAE for the held-out test data.

We used Python's \texttt{lifelines} package\footnote{\url{https://lifelines.readthedocs.io/en/latest/}} for an implementation of the Cox proportional hazards model and tuned over a range of L1 and L2 penalties, using the same training/validation/test splits. While \texttt{lifelines} has the basic implementation of a time-varying Cox model, it lacks the capacity to predict expected survival times on held-out data, being primarily used to report hazard ratios at different follow-up times. 

\section{Causal inference experiment}

We used an 80/20 training/validation split to tune DynST and the Cox model as outcome regression models, selecting for hyperparameters that produced the best validation MAE. We used the same hyperparameter spaces as in the prediction experiment. To fit the IPW estimator, we used the \texttt{sklearn} implementation of a cross-validated logistic regression with L2 penalty, fitted over 5 random folds. (Note that the standard deviation for the IPW estimator in Table 2 is zero; this is because a single model was validated over five random splits of the data.) This logistic model estimated propensity score as a function of three binary variables, indicating the presence of the diagnoses relevant to both treatment assignment and outcome (namely, hypertension, coronary atherosclerosis, and atrial fibrillation). Counterintuitively, an IPW estimator benefits from an underlying propensity model that uses a ``minimal'' set of confounding features that makes conservative estimates of treatment assignment (i.e., not too close to zero or one) \cite{aipw}. Using additional features to estimate propensity score did not improve ATE estimation.

We designed the treatment and outcome simulations so that the data exhibited a strong confounding structure: while the true effect of treatment is to \text{decrease} the hazards at all time steps, treatment assignment is strongly correlated with covariates that predict shorter survival times. Table \ref{true-ate} compares ATE (the true effect) and the unadjusted difference in means (the apparent effect) on RMST for three cutoff times. At each cutoff, treatment causes an increase in average survival time but correlates with a decrease in average survival time.

\begin{table}
  \caption{True ATEs versus unadjusted differences in means for RMST at varying cutoffs $\tau$}
  \label{true-ate}
  \centering
  \begin{tabular}{lll}
    \toprule
    $\tau$     & Average Treatment Effect    & Unadjusted Difference \\
    \midrule
    $8$ & $0.265 $ & $  -0.237 $ \\
    $12$ & $ 0.572$ & $ -0.539  $ \\
    $16$  &  $0.946$ & $-0.933$ \\
    \bottomrule
  \end{tabular}
\end{table}

\end{document}